%% file: depth-nerf.tex
\documentclass[letterpaper, 10 pt, conference]{ieeeconf}
\IEEEoverridecommandlockouts    
\input{stachnisslab-latex}


\usepackage{multirow}
\usepackage{vcell}
\usepackage{booktabs}
\usepackage{graphicx}
\usepackage[colorlinks,linkcolor=red,anchorcolor=blue,citecolor=green]{hyperref}
\usepackage{cite}
\usepackage{pifont}

\usepackage[capitalize]{cleveref}
\crefname{section}{Sec.}{Secs.}
\Crefname{section}{Section}{Sections}
\Crefname{table}{Table}{Tables}
\crefname{table}{Tab.}{Tabs.}
\usepackage[table,xcdraw]{xcolor}

\newcommand{\name}[0]{TD-NeRF}

\title{\LARGE \bf TD-NeRF: Novel Truncated Depth Prior for Joint Camera Pose and Neural Radiance Field Optimization} 

\author{Zhen Tan \and Zongtan Zhou \and Yangbing Ge \and Zi Wang \and Xieyuanli Chen$^*$  \and Dewen Hu$^*$
  \thanks{Z. Tan, Z. Zhou, Y. Ge, X. Chen, D. Hu are with the College of Intelligence Science and Technology, National University of Defense Technology, China. Z. Wang is with College of Aerospace Science and Engineering, National University of Defense Technology, China.}%
  \thanks{* indicates corresponding authors:  X. Chen (xieyuanli.chen@nudt.edu.cn) and D. Hu (dwhu@nudt.edu.cn)}
  \thanks{This work has partially been funded by the National Natural Science Foundation of China (12302252, U19A2083, 62403478), Young Elite Scientists Sponsorship Program by CAST (No. 2023QNRC001), and Key Programs for Science and Technology Development of Hunan (2024QK2006).}%
}

\begin{document}
\thispagestyle{empty}
\pagestyle{empty}
\maketitle

\begin{abstract}
The reliance on accurate camera poses is a significant barrier to the widespread deployment of Neural Radiance Fields (NeRF) models for 3D reconstruction and SLAM tasks. The existing method introduces monocular depth priors to jointly optimize the camera poses and NeRF,
which fails to fully exploit the depth priors and neglects the impact of their inherent noise.
In this paper, we propose Truncated Depth NeRF (TD-NeRF), a novel approach that enables training NeRF from unknown camera poses - by jointly optimizing learnable parameters of the radiance field and camera poses. 
Our approach explicitly utilizes monocular depth priors through three key advancements: 1) we propose a novel depth-based ray sampling strategy based on the truncated normal distribution, which improves the convergence speed and accuracy of pose estimation; 2) to circumvent local minima and refine depth geometry, we introduce a coarse-to-fine training strategy that progressively improves the depth precision; 3) we propose a more robust inter-frame point constraint that enhances robustness against depth noise during training.
The experimental results on three datasets demonstrate that \name{} achieves superior performance in the joint optimization of camera pose and NeRF, surpassing prior works, and generates more accurate depth geometry.
The implementation of our method has been released at \url{https://github.com/nubot-nudt/TD-NeRF}.
\end{abstract}

\section{Introduction}
\label{sec:intro}

Pose estimation and scene representation play crucial roles in 3D Reconstruction~\cite{schonberger2016structure, ozyecsil2017survey} and Simultaneous Localization and Mapping (SLAM)~\cite{bailey2006simultaneous}. In recent studies on scene representation methods~\cite{riegler2020free, tucker2020single, rusu20113d}, Neural Radiance Fields (NeRF)~\cite{mildenhall2021nerf} have gained significant attention in the domains of robotics and autonomous driving, primarily because of their capacity to produce highly realistic images. 

Given a set of posed images, 
NeRF~\cite{mildenhall2021nerf} is capable of simulating radiance fields using neural networks. 
Most current NeRF methods~\cite{barron2022mip, muller2022instant, roessle2022dense, jeong2021self} separate the pose estimation and reconstruction rendering processes.
They use offline processing methods like Structure from Motion (SfM)~\cite{schonberger2016structure, schoenberger2016mvs} to obtain camera poses from RGB images. 
Such images with poses are then fed into a radiance field network, known as posed NeRF. 
These loosely coupled approaches have multiple limitations. 
First, the accuracy of the rendering in NeRF relies on the accuracy of the camera poses. 
Second, the mutual correlation between the pose and the radiance field is ignored. In other words, camera poses can improve the accuracy of the radiance field, while radiance field information can assist in optimizing the camera pose. 
Third, failures occur when using methods such as COLMAP or SfM~\cite{schonberger2016structure, schoenberger2016mvs} to generate the poses, which makes it impossible to carry out subsequent work such as radiance field reconstruction, and there is no remedy. 
Fourth, the loosely coupled approach is challenging to deploy in real-world scenarios for 3D reconstruction and SLAM tasks with large motion changes.

To reduce the dependence on pose in NeRF, some methods~\cite{wang2021nerfmm, lin2021barf, jeong2021self, meng2021gnerf, bian2023nope} perform joint optimization of the poses and radiance fields.~\cite{wang2021nerfmm} adds a pose net for simultaneous optimization with the NeRF network, which is prone to fall into local optima or non-convergence;~\cite{lin2021barf} and~\cite{meng2021gnerf} use a priori information of the pose and refinement based on it, which is not effective when the pose is completely unknown;~\cite{jeong2021self} is optimized for different camera models and does not discuss the case where the poses are unknown. Furthermore, ~\cite{bian2023nope} introduces depth priors for supervision, which fails to fully exploit the depth priors and neglects the impact of their inherent noise.~\cite{deng2022depth} and ~\cite{roessle2022dense} also utilize depth priors, but these depth priors are derived from sparse point clouds computed through SfM and are not used for joint optimization.

\begin{figure}[t]
  \centering
  \includegraphics[width=\linewidth]{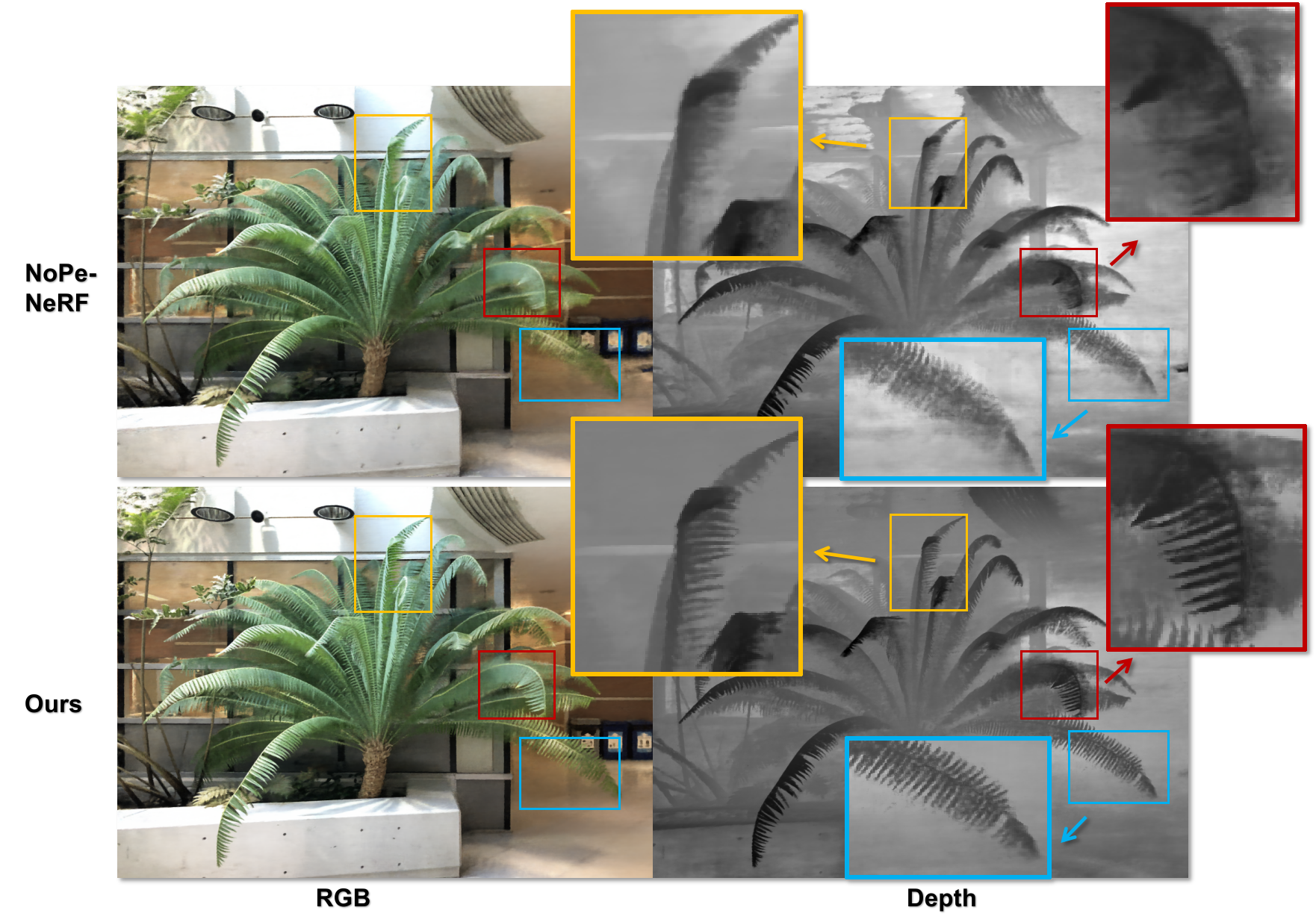}
  \caption{Comparison with the state-of-the-art depth-based NeRF method NoPe-NeRF~\cite{bian2023nope}. RGB images and depth images are rendered by NeRF with a coarse depth map.}
  \label{fig:motivation}
\end{figure}

In this paper, to jointly optimize camera poses and NeRF for enhanced pose estimation accuracy, we leverage depth priors derived from monocular depth estimation networks. First, we introduce a sampling strategy named Truncated Depth-Based Sampling (TDBS), which utilizes depth priors informed by the truncated normal distribution to optimize ray point sampling, as detailed in \cref{section:dbs}.
Specifically, 
we employ a coarse-to-fine training strategy in TDBS
to obtain more accurate depth geometry (see \cref{fig:motivation}). 
This strategy accelerates the convergence of pose optimization and enhances the precision of pose estimation. Second, we introduce a novel inter-frame point constraint (see \cref{section:gpc}). In this constraint, we utilize the Gaussian kernel function to measure distances between inter-frame point clouds, thereby robustly handling depth noise and improving the accuracy of relative pose estimation and the quality of view synthesis.

In summary, the contributions of this paper include:
(i) A robust depth-based NeRF method \name{} is proposed, which jointly optimizes camera poses and radiance fields based on depth priors. Our approach can be applied to both indoor and outdoor scenes with large motion changes.
(ii) We leverage depth priors for neural rendering and propose a truncated depth-based ray sampling strategy - TDBS. This strategy speeds up pose optimization convergence and improves the accuracy of pose estimation.
(iii) We further propose the coarse-to-fine training strategy for the sampling strategy, effectively mitigating local minima of the model and enhancing depth geometry resolution.
(iv) We propose a Gaussian point constraint that more robustly measures the distance between inter-frame point clouds, accounting for the depth noise.

\section{Related Work}
\label{sec:related}
\noindent\textbf{Visual SfM and SLAM:} Earlier work utilized SfM~\cite{schonberger2016structure, schoenberger2016mvs, pollefeys1999sfm} and SLAM~\cite{stachniss2016handbook-slamchapter, bailey2006simultaneous, mur2015orb} systems to simultaneously reconstruct 3D structures and estimate sensor poses. 
These methods could be categorized into indirect methods and direct methods, 
where indirect methods~\cite{davison2007monoslam, mur2015orb} employed keypoint detection and matching, 
while direct method~\cite{alismail2017photometric} utilized photometric error. 
However, they suffered from lighting variations and textureless scenes. 
To address this issue, NeRF~\cite{mildenhall2021nerf} represented 3D scenes using neural radiance fields and significantly improved the realism of scene representation. 
Given a set of posed images,
NeRF~\cite{mildenhall2021nerf} has demonstrated notable achievements in generating photo-realistic images and 3D reconstruction.
Furthermore, its variants~\cite{zhu2022nice, sucar2021imap, yen2021inerf} successfully combined NeRF with SLAM methods in indoor environments.

\noindent\textbf{NeRF and poses joint optimization:} 
Recent studies have started to focus on the pose-unknown NeRF. NeRFmm~\cite{wang2021nerfmm} jointly optimized the camera poses and the neural radiance network for forward-facing scenes but was susceptible to local optima. 
SC-NeRF~\cite{jeong2021self} proposed a joint optimization method that could be applied to different camera models. 
Inspired by SfM techniques, BARF~\cite{lin2021barf} and GARF~\cite{chng2022garf} introduced pose refinement methods. 
However, they relied on accurate initialization. 
To tackle this problem, GNeRF~\cite{meng2021gnerf} used Generative Adversarial Networks~\cite{creswell2018generative}, utilizing randomly initialized poses for complex outside-in scenarios. 
LU-NeRF~\cite{cheng2023lu} employed a connectivity graph-based approach for local-to-global pose estimation in 360° scenes. 
To achieve higher accuracy,~\cite{yen2021inerf, sucar2021imap} utilized depth sensor information to assist NeRF and joint pose optimization. 
More relevant to our work, NoPe-NeRF~\cite{bian2023nope} used a monocular depth network DPT~\cite{ranftl2021dpt} to estimate coarse depth maps and leveraged depth generation to enforce pose relative constraints.
However, it did not fully exploit the potential of monocular depth maps. 

Different from existing works, we reassess the utilization of depth priors and propose a coarse-to-fine ray sampling strategy in \cref{section:dbs}  to efficiently optimize poses while enhancing pose estimation and novel view synthesis accuracy.


\begin{figure*}[t]
  \centering
  \includegraphics[width=\linewidth]{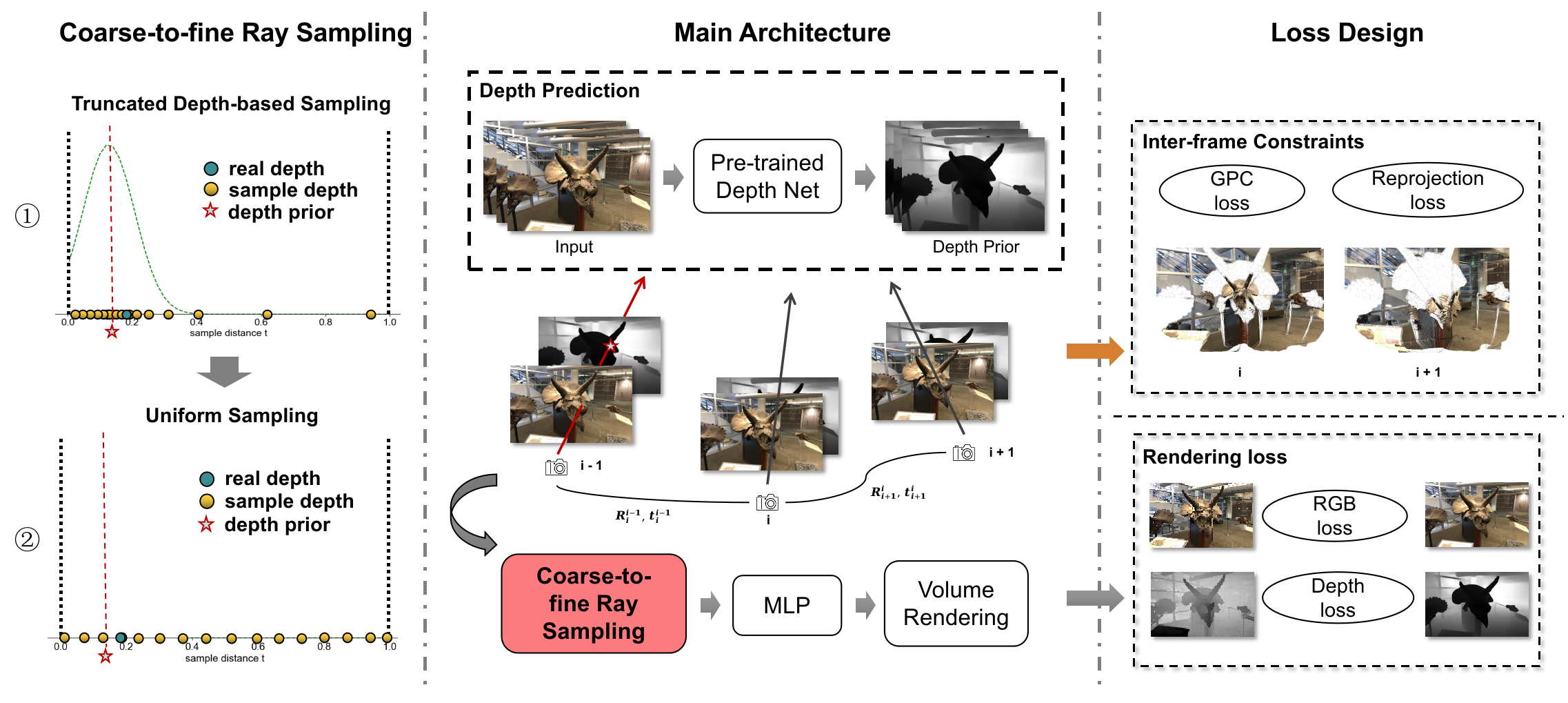}
  \caption{Overview. 
  The inputs are RGB images without poses, and RGB images are first processed by a pre-trained depth network to obtain depth priors. 
  Then, we employ a truncated normal distribution to optimize the ray sampling of each pixel based on the depth priors with a coarse-to-fine training strategy (\textcircled{\scriptsize{1}}: coarse step, \textcircled{\scriptsize{2}}: fine step).
  Subsequently, the sampled points are fed into an MLP to estimate the color $c$ and the density $\sigma$. 
  Next, RGB and depth images are integrated by color $c$ and $\sigma$ by utilizing volume rendering.
  Finally, the radiance field is optimized by supervising depth and RGB. 
  Additionally, we incorporate depth information to calculate GPC and reprojection loss between point clouds, providing constraints for inter-frame pose optimization and refinement.}
  \label{fig:overview}
  \vspace{-0.3cm}
\end{figure*}

\section{Method}
\label{sec:main}

The overview of our proposed method \name{} is illustrated in~\cref{fig:overview}. 
We tackle a crucial challenge in the current NeRF~\cite{mildenhall2021nerf} research, which involves simultaneously optimizing the neural radiance fields and camera pose without given camera poses in both indoor and outdoor scenes (\cref{sec:pre}). 
Firstly, we utilize a lightweight pre-trained depth estimation network, DPT~\cite{ranftl2021dpt} to obtain depth priors. To fully exploit depth priors and gain more accurate depth geometry, we propose a ray sampling strategy based on the truncated normal distribution~\cite{burkardt2014truncated} and depth priors called Truncated Depth-Based Sampling (TDBS). Further, to speed up the convergence during training, prevent falling into local minima, and make the rendered depth more accurate, we propose the coarse-to-fine training strategy for TDBS (\cref{section:dbs}).
Moreover, to enhance robustness against depth noise, we introduce a Gaussian Point Constraint (GPC) that measures the distance between inter-frame point clouds (\cref{section:gpc}.). 
Therefore, we effectively utilize the coarse depth map and seamlessly incorporate it into the synchronous training of NeRF and camera pose estimation.

\subsection{Preliminary}
\label{sec:pre}
\paragraph{NeRF~\cite{mildenhall2021nerf}}
It essentially models a mapping function that utilizes Multi-Layer Perceptron (MLP) layers, denoted as $F_{\theta}$, which maps 3D points $x$ and viewing directions $\mathbf{d}$ to color $c$ and volume density $\sigma$, i.e., $F_{\theta}(x, d) \xrightarrow{} (c, \sigma)$. 
The rendering process can be described as follows: 
1) Given a sequence of images, the camera ray $\mathbf{r}(t)=\mathbf{o}+t \mathbf{d}$ is casting into the scene, and a set of points $x$ is sampled along the ray; 
2) The network $F_{\theta}$ is employed to estimate the density $\sigma$ and color $c$ of this set of points; 
3) Volume rendering is used to integrate along the ray and obtain the color values $C(\mathbf{r})$ for synthesizing the final image. The entire integration process between $[t_n, t_f]$ is modeled as:
\begin{equation}
C(\mathbf{r})=\int_{t_n}^{t_f} T(t) \sigma(\mathbf{r}(t)) c(\mathbf{r}(t), \mathbf{d}) d t,
\end{equation}
where $T(t)=\exp \left(-\int_{t_n}^t \sigma(\mathbf{r}(s)) d s\right)$ is accumulated transmittance along a ray. More details could be referred to in~\cite{mildenhall2021nerf}.

\paragraph{Joint Optimization of NeRF and Poses} 
In the problem of pose-unknown NeRF, we simultaneously optimize the pose $P$ and parameters $\theta$ of NeRF. In existing methods, most approaches directly minimize the photometric error to obtain the final result. The mathematical expression is as follows:
\begin{equation}
\underset{P, \theta}{\operatorname{argmin}} \, d(C(\mathbf{r}), \hat{C}(\mathbf{r})) .
\end{equation}
In NoPe-NeRF~\cite{bian2023nope}, $\mathcal{L}_{self-depth}$, $\mathcal{L}_{pc}$, and $\mathcal{L}_{rgb-s}$ are defined by incorporating the depth map to self-supervise depth and constrain inter-frame variations.
The integration of the depth map ${D}^{nerf}_i(\mathbf{r})$ can be achieved through point sampling, similar to the volume integration process for color. Mathematically, this can be expressed as follows:
\begin{equation}\label{eq:depth_inte}
    {D}^{nerf}_i(\mathbf{r})=\int_{t_n}^{t_f} T(t) \sigma(\mathbf{r}(t)) d t ,
\end{equation}
where $t_n$ and $t_f$ denote the farthest and nearest distances of the sampling points, respectively.
Further, we use a self-supervised method to compute the loss between the rendered depth map ${D}^{nerf}_i(\mathbf{r})$ and the priori depth map $\hat{D}_i^{dpt}$. Since the a priori depth map obtained by the pre-trained monocular depth network does not have multi-view consistency, to recover a sequence of multi-view consistent depth maps, we use a linear transformation $s_i\hat{D}_i^{dpt}+k_i$ to undistort depth maps:
\begin{equation}
\label{eq:self-depth}
    \mathcal{L}_{\text {self-depth }}=\sum_i^N\left\|(s_i\hat{D}_i^{dpt}+k_i)-D^{nerf}_i\right\| .
\end{equation}
where $\{(s_i, k_i) | i=0,...,N\}$ are learnable parameters that will be optimized along with the NeRF network.
From \cref{eq:depth_inte}, the sampling points are sampled uniformly in that interval $[t_n, t_f]$, however,  this ignores the role of the depth priors.
In the following \cref{section:dbs}, we leverage the depth priors and improve the integration of $D^{nerf}_i(\mathbf{r})$  by proposing TDBS to make the depth self-supervising more robust.

\subsection{Truncated Depth-Based Sampling (TDBS)}
\label{section:dbs}
\paragraph{full-stage} In our approach, we utilize a monocular depth network, DPT~\cite{ranftl2021dpt}, to estimate the depth of each image. 
However, the depth map estimated by the pre-trained network is a coarse depth map, where the depth is generally accurate in most regions but fails to capture the fine details of certain objects. The previous state-of-the-art method, Nope-NeRF~\cite{bian2023nope}, considers overall depth distortion, which does not fully exploit the depth information. 
Therefore, we rethink the role of the monocular depth priors in rendering and propose that a coarse depth map can provide a prior for ray sampling, assisting in sampling by assuming that the real surface is near the estimated depth. 
To achieve this goal, we design the TDBS ray sampling strategy based on the truncated normal distribution~\cite{burkardt2014truncated}. This strategy enables sampling from a truncated normal distribution $\psi(\bar{\mu}, \bar{\sigma}, a, b ; x)$ with mean $\bar{\mu}$  (depth value) and variance $\bar{\sigma}$, truncated within the interval $[a, b]$. The mathematical expression of the probability density function $\psi(\cdot)$ is as follows:
\begin{equation}
\psi(\bar{\mu}, \bar{\sigma}, a, b ; x)=\left\{\begin{array}{cc}
0 & x \leq a ; \\
\frac{\phi\left(\bar{\mu}, \bar{\sigma}^2 ; x\right)}{\Phi\left(\bar{\mu}, \bar{\sigma}^2 ; b\right)-\Phi\left(\bar{\mu}, \bar{\sigma}^2 ; a\right)} & a<x<b ; \\
0 & b \leq x ,
\end{array}\right.
\end{equation}
where $\phi(\cdot)$ and $\Phi(\cdot)$ denote the probability density and distribution function of the standard normal distribution, respectively. 
We use inverse distribution sampling~\cite{mosegaard1995monte} to obtain sampling points along the ray. 
The number of sampling points is set to 128. In practice, the values of $a$, $b$, and $\sigma$ are set to 0.001, 1.0, and 0.1, respectively.

\paragraph{Coarse-to-fine Training Strategy} Relying on the aforementioned strategy, we can predict the depth of the majority of pixels with relative accuracy. However, the depth estimated by DPT is inaccurate or even incorrect in many details. 
If we use full-stage TDBS, these erroneous depth values tend to degrade the rendering quality. Therefore, to avoid the model from falling into a local minima during the training, we propose a coarse-to-fine training strategy for TDBS. 
In the coarse stage, we utilize TDBS to converge the density values of the majority of points to the optimal values. 
In the fine stage, 
we adopt Uniform Sampling.
The purpose of this stage is to recompute the erroneous depth values based on the already optimized majority of points, aiming to optimize the noisy and erroneous points. 
In summary, the coarse-to-fine TDBS strategy can be defined as follows:
\begin{equation}
    S(\mathbf{r})= \begin{cases} S_\text{TDBS}(\mathbf{r}) & \text { epoch }<T_s \\ 
        S_{\text {uni}}(\mathbf{r}) & \text { epoch } \geqslant T_s \end{cases} , 
\end{equation}
where $S_\text{TDBS}(\mathbf{r})$ and $S_{\text{uni}}(\mathbf{r})$ denote TDBS and Uniform Sampling, respectively. $T_s$ is a threshold controlling the sampling method. In our approach, $T_s$ corresponds to the epoch at which the learning rate begins to schedule.
In this way, according to \cref{eq:self-depth}, a more robust and accurate initial value of $D^{nerf}_i(\mathbf{r})$ can be obtained, which is self-supervised with $D^{dpt}_i(\mathbf{r})$, resulting in a fast and robust convergence of the model. We empirically set $T_s$ to 1000.

\subsection{Gaussian Point Constraint (GPC)}
\label{section:gpc}
Given the depth point clouds of a scene, a method~\cite{bian2023nope} utilizes the Chamfer distance~\cite{butt1998chamfer} to measure the distance between the inter-frame point clouds as a constraint for pose estimation. 
However, the Chamfer distance can be sensitive to noise in the estimated depth map, since it imposes a strong constraint on point clouds and can penalize even small deviations between the estimated and ground truth point clouds.

Therefore, we propose a point cloud constraint based on the Gaussian kernel function~\cite{keerthi2003gaussian}, 
which is expressed as:
\begin{equation}
w_{i j}=\exp \left(-\frac{\left\|\mathbf{p}_i^m-\mathbf{p}_j^{m+1}\right\|^2}{2 \sigma_{p c}^2}\right) ,
\end{equation}
where $w_{ij}$ denotes the weight between points $\mathbf{p}^m_i$ and $\mathbf{p}^{m+1}_j$, $\mathbf{p}^m_i$ represents the $i_{th}$ point in the $m_{th}$ frame of the point cloud, $\sigma_{pc}$ denotes the standard deviation of the Gaussian kernel function, and ${\| \mathbf{p}_i - \mathbf{p}_j \|^2}$ represents the Euclidean distance between points $\mathbf{p}_i$ and $\mathbf{p}_j$.
Here, $\sigma_{pc}$ is empirically set to 1.

By applying the Gaussian kernel function to the distance matrix, the corresponding weight values can be determined based on the proximity of points. 
Points that are closer together have higher weights, while points that are farther apart have lower weights. 
This allows for greater emphasis on the distances between adjacent points, thus more accurately reflecting the relationships between point clouds. 
We can incorporate this into the regularization term $\mathcal{L}_{GPC}$ :
\begin{equation}
D_{i j}=w_{i j} \cdot\left\|\mathbf{p}_i^m-\mathbf{p}_j^{m+1}\right\| ,
\end{equation}
\begin{equation}
\mathcal{L}_{GPC}=\sum D_{i j} ,
\end{equation}
where $D_{ij}$ is the distance between points $\mathbf{p}^m_i$ and $\mathbf{p}^{m+1}_j$.

\subsection{Reprojection Loss} 
The GPC mentioned above establishes associations between 3D points across frames. 
Additionally, the photometric error between pixel correspondences across frames is also minimized. 
Hence, the reprojected photometric loss is defined as:
\begin{equation}
    \mathcal{L}_{reproj}= \sum_{(m,n)}\left\|I_m\left(K_m P_m\right),  I_{n}\left(K_{n}\left(R_m^{n} P_{m}+t_m^{n}\right)\right\|\right . ,
\end{equation}
where $(R^n_m, t^n_m)$ denote the rotation matrix and translation from the $m_{th}$ camera coordinate to the $n_{th}$ camera coordinate, 
$P_m$ represents the point cloud derived from the $m_{th}$ depth map and RGB image, 
$K_m$ denotes a projection matrix for the $m_{th}$ camera, 
and $I(\cdot)$ represents the pixel value of the point in the image.

\subsection{Overall Training Loss}
Assembling all loss terms, we get the overall loss function:
\begin{equation}\label{eq:overall-loss}
\mathcal{L}=\mathcal{L}_{r g b}+\lambda_1 \mathcal{L}_{self-depth}+\lambda_2 \mathcal{L}_{GPC}+\lambda_3 \mathcal{L}_{reproj} ,
\end{equation}
where $\lambda_1$, $\lambda_2$, $\lambda_3$ represent weights assigned to different loss terms. In practice, we set $\lambda_1 = 0.04$, $\lambda_2 = 1.0$, $\lambda_3 = 1.0$.


\section{Experimental Evaluation}
\label{sec:exp}
%
The main focus of this work is how the depth prior can be leveraged to achieve simultaneous optimization of NeRF and camera pose. 
We compare our method with pose-unknown methods: Nope-NeRF~\cite{bian2023nope} and NeRFmm~\cite{wang2021nerfmm}.

\begin{figure*}[t]
  \centering
  \includegraphics[width=0.98\linewidth, height=0.82\linewidth]{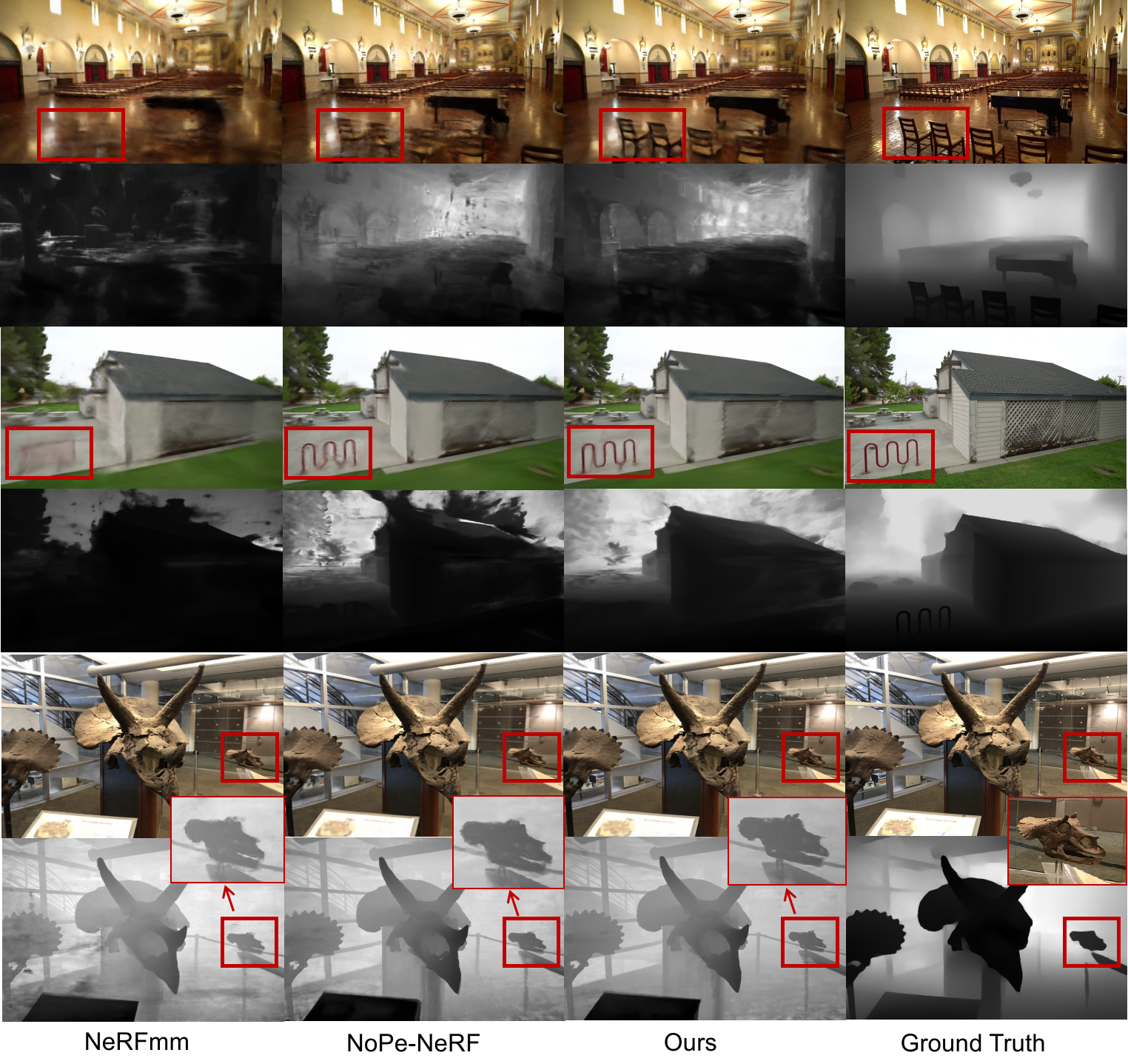}
  \caption{Qualitative Comparison of Novel View Synthesis on Tanks \& Temples (top: 4 rows) and LLFF (bottom: last 2 rows) dataset.  The rendered RGB and depth images are visualized above. \name{} is able to recover better details for both RGB and depth geometry, as shown in the red box. (The ground truth of depth is generated by DPT~\cite{ranftl2021dpt}.)}
  \label{fig:nvs-tanks}
\end{figure*}

We present our experiments to show the capabilities of our method. The results of our experiments also support our key claims, which are:
(i) Our proposed method \name{} can optimize the camera pose and radiance field simultaneously based on the depth prior and can be applied to both indoor and outdoor scenes. A specific large motion changes dataset validates that our method can be applied in the presence of large motion changes as well.
(ii) The proposed TDBS sampling strategy based on the depth prior not only speeds up the convergence of training but also yields better pose estimation results.
(iii) Further, the experimental results verify that our proposed coarse-to-fine training strategy approach is effective in avoiding falling into local optima and obtaining more accurate depth geometry.
(iv) Finally, the experiments verify that the Gaussian point constraint is more robust compared to other inter-frame point constraints.

\subsection{Experimental Setup}

\paragraph{Datasets} 
We conduct experiments on three datasets
to evaluate the performance of our method,
including LLFF~\cite{mildenhall2019llff}, Tanks and Temples~\cite{knapitsch2017tanks}, and BLEFF~\cite{wang2021nerfmm}. 
\newline\textbf{LLFF}~\cite{mildenhall2019llff}: We first conduct experiments on the forward-facing dataset as that in NeRFmm~\cite{wang2021nerfmm} and NeRF~\cite{mildenhall2021nerf}, which has 8 scenes containing 20-62 images. The resolution of the training images is 756 $\times$ 1008, and every 8th image is used for novel view synthesis. 
\newline\textbf{Tanks and Temples}~\cite{knapitsch2017tanks}: 8 scenes are used to evaluate the quality of novel view synthesis and camera pose estimation. To emulate scenarios of bandwidth limitations in data transmission and large motion changes, we employed a sampling approach on the dataset provided by Nope-NeRF~\cite{bian2023nope}. Specifically, we selected every 5th image, resulting in a reduced frame rate of 6 fps (origin: 30 fps). During training, we utilized images with a resolution of 960 $\times$ 540. Similarly, for the Family scene, we divided the frames equally, allocating half for training and the remaining half for testing.
\newline\textbf{BLEFF}~\cite{wang2021nerfmm}: This dataset proposed by NeRFmm has the ground truth of the pose and can be used specifically to check the accuracy of the pose estimation. To simulate the case of large motion changes, we select 5 scenes of subset $t_{010}r_{010}$ with large rotation and translation to evaluate pose accuracy and novel view synthesis. Each scene consists of 31 images, all of which are sampled to a resolution of 780 $\times$ 520.

\paragraph{Metrics} 
To evaluate our proposed method, 
we consider two aspects of test views, 
including the quality of novel view synthesis and camera pose estimation. 
\newline\textbf{For novel view synthesis:} following previous methods~\cite{mildenhall2021nerf, barron2022mip, muller2022instant}, we use (1) Peak signal-to-noise ratio (PSNR); (2) Structural Similarity Index (SSIM); and (3) Learned Perceptual Image Patch Similarity (LPIPS). 
\newline\textbf{For camera pose evaluation:} we use visual odometry metrics~\cite{nister2004vo}; (4) Relative Pose Error (RPE); and (5) Absolute Trajectory Error (ATE).


\paragraph{Implementation Details} We implement our approach in PyTorch, Ubuntu20.04, RTX4090. We adopt the multi-network structure based on NoPe-NeRF~\cite{bian2023nope} without any modification. The activation function is Softplus. We sample 128 points along each ray using a truncated normal distribution combined with depth priors (\cref{section:dbs}) with noise within a predefined range. 
For LLFF, the predefined range is (0.0, 1.0), and for Tanks and Temples, the range is (0.1, 10). 
The initial learning rate for NeRF is 0.001 and for pose and distortion is 0.0005. 
The training process is divided into 3 phases: the first phase is the simultaneous optimization of all losses; the second phase gradually reduces all weights to 0 except RGB loss; and in the third phase, the network is finely optimized for a single loss. All models are trained for 12000 epochs unless otherwise specified. The initial weights of the different losses are $\lambda_{1}=0.04, \lambda_{2}=1.0, \lambda_{3}=1.0$.

\subsection{Performance}

\subsubsection{On Camera Pose Estimation }
\begin{table}[t]
\caption{Camera pose estimation results on LLFF.}
\centering
\scriptsize
\renewcommand\arraystretch{1.1}
\setlength{\tabcolsep}{3.4pt}
\label{tab:pose-llff}
\begin{tabular}{@{}lccccccccc@{}}
\toprule
& \multicolumn{3}{c}{Ours} & \multicolumn{3}{c}{Nope-NeRF} & \multicolumn{3}{c}{NeRFmm} \\ \cmidrule(lr){2-4} \cmidrule(lr){5-7} \cmidrule(lr){8-10} 
\multirow{-2}{*}{Scenes}  &  RPE$_t$  &  RPE$_r$ &  ATE  &  RPE$_t$  &  RPE$_r$ &  ATE  &  RPE$_t$  &  RPE$_r$  &  ATE \\ \midrule 
fern & \textbf{0.088} & \textbf{0.1413} & \textbf{0.0008} & 0.326 & 1.2690 & 0.0040 & \textbackslash{} & 1.780 & 0.0290  \\
flower & \textbf{0.036} & \textbf{0.0482} & \textbf{0.0005} & 0.055 & 0.2361 & 0.0009 & \textbackslash{} & 4.840 & 0.0160  \\
fortress & \textbf{0.066} & \textbf{0.2310} & \textbf{0.0010} & 0.104 & 0.4550 & 0.0020 & \textbackslash{} & 1.360 & 0.0250  \\
horns & \textbf{0.177} & \textbf{0.3320} & \textbf{0.0030} & 0.264 & 0.5440 & 0.0050 & \textbackslash{} & 5.550 & 0.0440  \\
leaves & \textbf{0.140} & \textbf{0.0270} & \textbf{0.0013} & 0.147 & 0.0474 & 0.0014 & \textbackslash{} & 3.900 & 0.0160  \\
orchids & \textbf{0.135} & \textbf{0.1121} & \textbf{0.0016} & 0.313 & 0.9708 & 0.0051 & \textbackslash{} & 4.960 & 0.0510  \\
room & \textbf{0.082} & \textbf{0.2846} & \textbf{0.0014} & 0.329 & 1.0339 & 0.0054 & \textbackslash{} & 2.770 & 0.0300  \\
trex & \textbf{0.432} & \textbf{0.6034} & \textbf{0.0068} & 0.557 & 0.7339 & 0.0089 & \textbackslash{} & 4.670 & 0.0360  \\ \midrule 
average & \textbf{0.145} & \textbf{0.2225} & \textbf{0.0021} & 0.262 & 0.6613 & 0.0041 & \textbackslash{} & 3.729 & 0.0309  \\ \bottomrule
\end{tabular}%
\end{table}

\begin{table}[]
\renewcommand\arraystretch{1}
\setlength{\tabcolsep}{7pt}
\scriptsize
\centering
\caption{Quantitative evaluation of novel view synthesis and camera pose estimation on Tanks and Temples (averaged over 8 scenes). We adopt the COLMAP pose as the ground truth. 
Unlike the dataset provided by NoPe-NeRF~\cite{bian2023nope}, 
to simulate the \textbf{bandwidth-constrained scenario}, the original data is sampled at 1 frame every 5 frames, and the frequency is reduced from 30 fps to 6 fps. ( \dag : indicates our re-implementation in 6 fps.) }
\label{tab:nvs-tanks}
\begin{tabular}{@{}lcccccc@{}}
\toprule
     & \multicolumn{3}{c}{View Synthesis Quality}     & \multicolumn{3}{c}{Camera Pose Estimation}          \\ \cmidrule(lr){2-4} \cmidrule(lr){5-7} 
          & PSNR$\uparrow$  & SSIM$\uparrow$ & LPIPS$\downarrow$ & RPE$_t$ & RPE$_r$ & ATE    \\ \midrule
NeRFmm$\dagger$ & 21.26 & 0.57 & 0.56  & 6.6800 & 1.6345 & 0.1663 \\
Nope-NeRF$\dagger$ & 23.63 & 0.67 & 0.46  & 3.6065 & 0.4991 & 0.0909 \\
ours & \textbf{23.76} & \textbf{0.67} & \textbf{0.45} & \textbf{3.1759} & \textbf{0.4728} & \textbf{0.0762} \\ \bottomrule
\end{tabular}
\end{table}
\begin{table}[t]
\renewcommand\arraystretch{1.1}
\setlength{\tabcolsep}{1.8pt}
\scriptsize
\centering
\caption{Quantitative evaluation of camera pose estimation on BLEFF (subset: $t_{010}r_{010}$). The ground truth of the pose is provided by BLEFF dataset.}
\label{tab:pose-bleff}
\begin{tabular}{@{}lccccccccc@{}}
\toprule
\multirow{2}{*}{Scenes} & \multicolumn{3}{c}{Ours}                             & \multicolumn{3}{c}{Nope-NeRF} & \multicolumn{3}{c}{NeRFmm}   \\ \cmidrule(lr){2-4} \cmidrule(lr){5-7} \cmidrule(lr){8-10} 
                        & RPE$_t$         & RPE$_r$       & ATE            & RPE$_t$  & RPE$_r$ & ATE  & RPE$_t$ & RPE$_r$ & ATE   \\ \midrule
airplane               & \textbf{0.2786} & \textbf{0.0904}  & \textbf{0.0028} & 23.0097  & 2.5415    & 0.3661 & 0.3206  & 0.1004    & 0.0029 \\
bed                    & \textbf{0.2650} & \textbf{0.02391} & \textbf{0.0025} & 0.9045   & 0.1023    & 0.0091 & 0.2709  & 0.0246    & 0.0025 \\
classroom              & \textbf{0.4783} & \textbf{0.0479}  & \textbf{0.0053} & 1.9869   & 0.2140    & 0.0177 & 1.3417  & 0.1324    & 0.0115 \\
halloween & \textbf{0.2691} & \textbf{0.1165} & \textbf{0.0028} & 5.6785 & 1.2731 & 0.1140 & 10.7105 & 1.0577 & 0.2024 \\
castle                  & \textbf{0.4380} & \textbf{0.0943}  & \textbf{0.0038} & 1.4011   & 0.3094    & 0.0123 & 0.5669  & 0.1220    & 0.0050 \\ \midrule
average                 & \textbf{0.3458} & \textbf{0.0746}  & \textbf{0.0034} & 6.5961   & 0.8881    & 0.1038 & 2.6421  & 0.2874    & 0.0449 \\ \bottomrule
\end{tabular}%
\end{table}

\begin{figure}[t]
  \centering
  \includegraphics[width=\columnwidth,  height=0.8\linewidth]{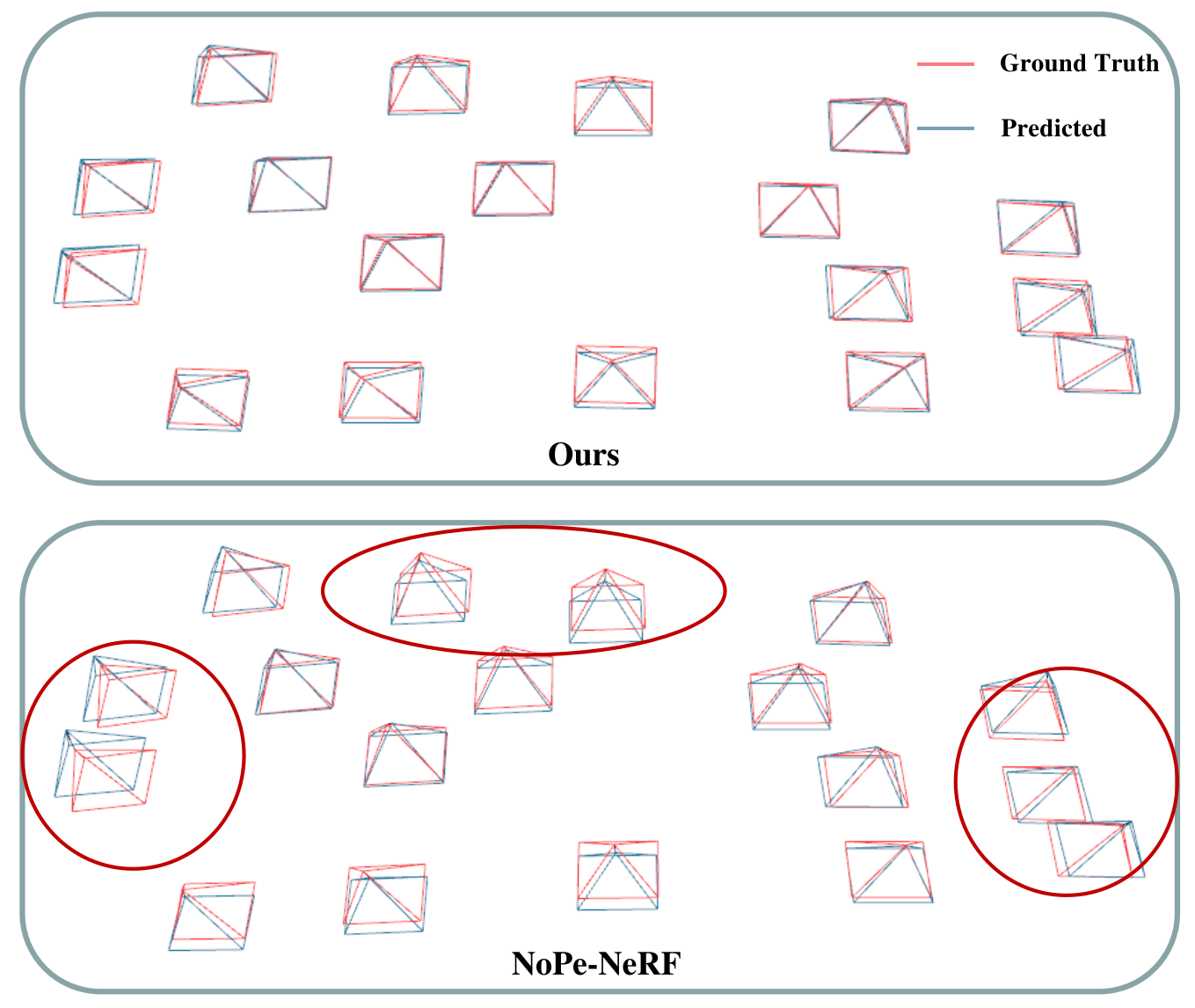}
  \caption{Pose Estimation Comparison. We visualize the camera poses on LLFF (scene: fern). red: ground truth; blue: predicted pose}
  \label{fig:pose-vis}
  \vspace{-0.2cm}
\end{figure}

Our proposed method significantly outperforms other baselines in all metrics for camera pose estimation. 
NoPe-NeRF~\cite{bian2023nope}, a state-of-the-art NeRF designed for pose-unknown novel view synthesis, and NeRFmm~\cite{wang2021nerfmm}, a key benchmark in pose-unknown NeRF approaches, were evaluated on three datasets. 
For LLFF~\cite{mildenhall2019llff} and Tanks and Temples~\cite{knapitsch2017tanks}, where the ground truth pose is not available, we utilized the pose provided by COLMAP~\cite{pollefeys1999sfm} as the reference. As shown in \cref{tab:pose-llff,tab:nvs-tanks}. Our method demonstrates a significant reduction in errors on the LLFF, achieving improvements of 44.8\%, 66.4\%, and 49.8\% compared to the state-of-the-art methods. Similarly, on the Tanks and Temples, our method achieves reductions in errors of 8.88\%, 5.27\%, and 10.02\%, respectively. To further validate the robustness of our approach, we conducted experiments on the BLEFF~\cite{wang2021nerfmm}, which contains the ground truth camera poses. Our method outperforms the current state-of-the-art method by a considerable margin in terms of error reduction. We provide quantitative results and a qualitative visualization in \cref{tab:pose-bleff} and \cref{fig:pose-vis}, respectively. 
Our method achieves better quantitative and qualitative results than previous state-of-the-art methods. This experiment demonstrates that our method can optimize the camera pose simultaneously in both indoor and outdoor scenes with large motion changes.

\subsubsection{On Novel View Synthesis}

\begin{table}[t]
\renewcommand\arraystretch{1.1}
\setlength{\tabcolsep}{0.65pt}
\scriptsize
\centering
\caption{Novel View Synthesis results on LLFF dataset.}
\label{tab:nvs-llff}
\begin{tabular}{@{}lcccccccccccc@{}}
\toprule
 & \multicolumn{3}{c}{Ours} & \multicolumn{3}{c}{Nope-NeRF} & \multicolumn{3}{c}{NeRFmm} & \multicolumn{3}{c}{COLMAP} \\ \cmidrule(lr){2-4} \cmidrule(lr){5-7} \cmidrule(lr){8-10} \cmidrule(lr){11-13} 
\multirow{-2}{*}{Scenes} &  PSNR  & SSIM  & LPIPS & PSNR & SSIM & LPIPS & PSNR & SSIM & LPIPS & PSNR & SSIM & LPIPS \\ \midrule \midrule 
fern & 20.08 & 0.61 & \textbf{0.40} & {\color[HTML]{333333} 19.79} & {\color[HTML]{333333} 0.59} & {\color[HTML]{333333} 0.44} & 21.67 & 0.61 & 0.50 & \textbf{22.22} & \textbf{0.64} & 0.47 \\
flower & 28.88 & \textbf{0.85} & \textbf{0.19} & \textbf{29.04} & \textbf{0.85} & \textbf{0.19} & 25.34 & 0.71 & 0.37 & 25.25 & 0.71 & 0.36 \\
fortress & \textbf{28.67} & \textbf{0.79} & \textbf{0.25} & 27.46 & 0.74 & 0.26 & 26.20 & 0.63 & 0.49 & 27.60 & 0.73 & 0.38 \\
horns & \textbf{25.80} & \textbf{0.76} & \textbf{0.34} & 25.00 & 0.73 & 0.37 & 22.53 & 0.61 & 0.50 & 24.25 & 0.68 & 0.44 \\
leaves & \textbf{20.23} & \textbf{0.64} & \textbf{0.37} & 19.97 & 0.63 & 0.38 & 18.88 & 0.53 & 0.47 & 18.81 & 0.52 & 0.47 \\
orchids & 17.90 & 0.51 & \textbf{0.41} & 18.03 & 0.49 & 0.43 & 16.73 & \textbf{0.55} & 0.39 & \textbf{19.09} & 0.51 & 0.46 \\
room & \textbf{28.02} & \textbf{0.90} & \textbf{0.26} & 27.90 & 0.89 & 0.29 & 25.84 & 0.84 & 0.44 & 27.77 & 0.87 & 0.40 \\
trex & \textbf{25.40} & \textbf{0.83} & \textbf{0.30} & 24.89 & 0.81 & 0.32 & 22.67 & 0.72 & 0.44 & 23.19 & 0.74 & 0.41 \\ \midrule
average & \textbf{24.37} & \textbf{0.74} & \textbf{0.32} & 24.01 & 0.72 & 0.34 & 22.48 & 0.65 & 0.45 & 23.52 & 0.68 & 0.42 \\ \bottomrule
\end{tabular}%
\end{table}
\begin{figure*}[]
  \centering
  \includegraphics[width=0.9\textwidth]{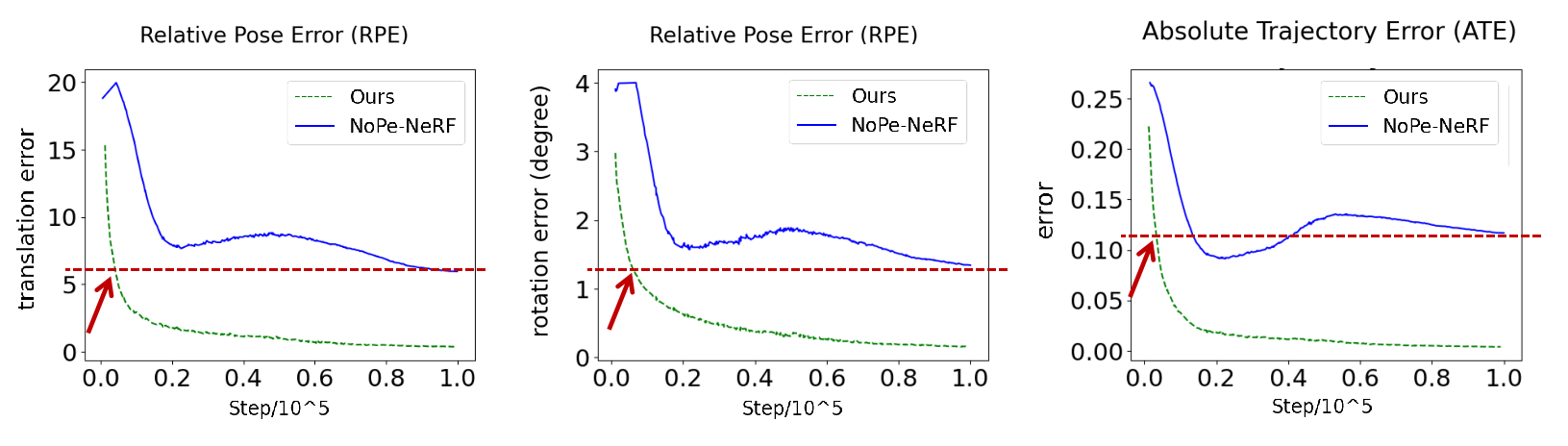}
  \vspace{-0.32cm}
  \caption{Visualization of convergence. 
  The experiment is conducted on the dataset BLEFF (scene: bed1).
  The blue and green colors denote NoPe-NeRF and ours, respectively. At the red arrow, the error of our method already reaches the final convergence result of Nope-NeRF.}
  \label{fig:opt-time}
\end{figure*}

 \begin{figure}[t]
  \centering
  \includegraphics[width=\columnwidth]{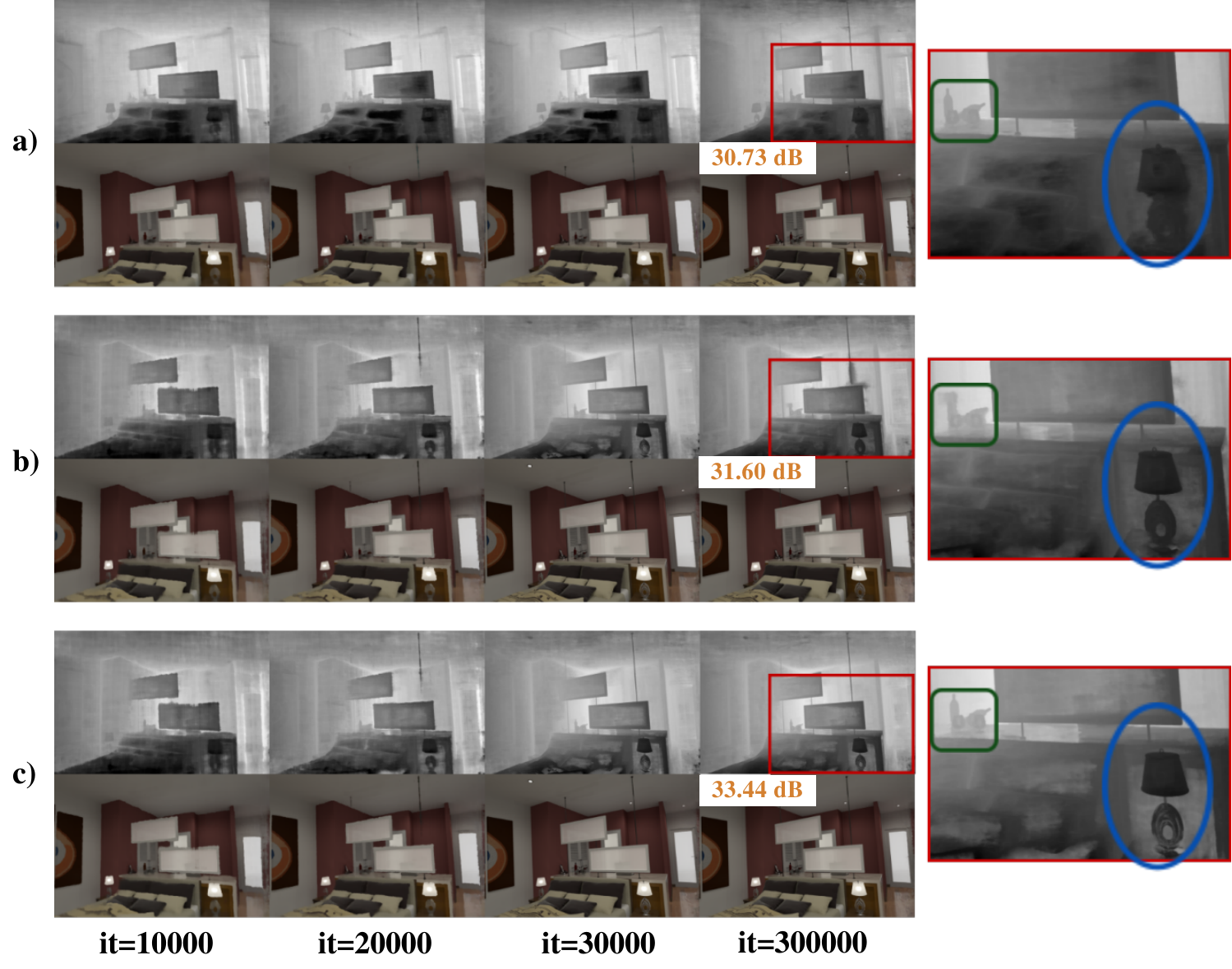}
  \vspace{-0.5cm}
  \caption{Visual comparison of TDBS strategy over training time. We compare a baseline method with two different training strategies on the BLEFF dataset (scene: bed1). a: NoPe-NeRF~\cite{bian2023nope}, b: the full-stage TDBS, and c: the coarse-to-fine TDBS.}
  \label{fig:TDBS}
\end{figure}

The second experiment evaluates the quality of novel view synthesis and illustrates that our approach is capable of further improving the quality of rendering on different datasets. We visualize the results on Tanks \& Temples and LLFF datasets, as shown in \cref{fig:nvs-tanks}.  Our method can better reproduce the rendering details in both indoor and outdoor scenes. Especially in outdoor scenes, our method renders depth geometry well. Whereas previous methods tend to perform poorly in such open scenes. The quantitative results are summarised in \cref{tab:nvs-tanks} and \cref{tab:nvs-llff}.

\begin{table}[]
\renewcommand\arraystretch{1}
\setlength{\tabcolsep}{5.5pt}
\scriptsize
\centering
\caption{Ablation study results of different training strategies on BLEFF (scene: bed1)}
\label{tab:bleff-tdbs}
\begin{tabular}{@{}lcccccc@{}}
\toprule
     & \multicolumn{3}{c}{View Synthesis Quality}     & \multicolumn{3}{c}{Camera Pose Estimation}          \\ \cmidrule(lr){2-4} \cmidrule(lr){5-7} 
          & PSNR$\uparrow$  & SSIM$\uparrow$ & LPIPS$\downarrow$ & RPE$_t$ & RPE$_r$ & ATE    \\ \midrule
uniform& 30.73& \textbf{0.95}& 0.12
& 0.9045& 0.1023& 0.0091 
\\
full-step TDBS& 31.60 
& 0.94& 0.14
& 0.4900& 0.0487& 0.0045
\\
coarse-to-fine TDBS& \textbf{33.44}& \textbf{0.95}& \textbf{0.11}& \textbf{0.2650}& \textbf{0.0239}& \textbf{0.0025}\\ \bottomrule
\end{tabular}
\vspace{-0.2cm}
\end{table}

\begin{table}[t]
\renewcommand\arraystretch{1}
\setlength{\tabcolsep}{4.5pt}
\scriptsize
\centering
\caption{Ablation study results of different strategies on LLFF (scene: fern).}
\label{tab:ablation}
\begin{tabular}{@{}lcccccc@{}}
\toprule
     & \multicolumn{3}{c}{View Synthesis Quality}     & \multicolumn{3}{c}{Camera Pose Estimation}         \\ \cmidrule(lr){2-4} \cmidrule(lr){5-7} 
                     & PSNR$\uparrow$  & SSIM$\uparrow$& LPIPS$\downarrow$& RPE$_t$ & RPE$_r$ & ATE    \\ \midrule
ours w/o  GPC         & 20.16 & 0.60 & 0.42  & 0.092  & 0.1780 & \textbf{0.0009} \\
ours w/o TDBS        & 20.08 & 0.61 & 0.40  & 0.125  & 0.3650 & 0.0010 \\
ours w/o TDBS + GPC  & 19.99 & 0.60 & 0.42  & 0.207  & 0.7643 & 0.0024 \\ \midrule
baseline (NoPe-NeRF) & 19.79 & 0.59 & 0.44  & 0.326  & 1.2690 & 0.0040 \\ 
ours & \textbf{20.20} & \textbf{0.61} & \textbf{0.39} & \textbf{0.092} & \textbf{0.1450} & \textbf{0.0009} \\ \bottomrule
\end{tabular}%
\vspace{0.4cm}
\renewcommand\arraystretch{1}
\setlength{\tabcolsep}{7pt}
\scriptsize
\centering
\caption{Ablation study results of different point constraints on LLFF (scene: fern).}
\label{tab:ablation-point-constraint}
\begin{tabular}{@{}lcccccc@{}}
\toprule
     & \multicolumn{3}{c}{View Synthesis Quality}     & \multicolumn{3}{c}{Camera Pose Estimation}         \\ \cmidrule(lr){2-4} \cmidrule(lr){5-7} 
                     & PSNR$\uparrow$  & SSIM$\uparrow$& LPIPS$\downarrow$& RPE$_t$ & RPE$_r$ & ATE    \\ \midrule
distribution& 19.75& 0.58& 0.45& 0.398& 1.3081& 0.0040\\
KL divergence& 19.95
& 0.59& 0.43& 0.251& 0.9587& 0.0029
\\
 ICP& 19.79& 0.59& 0.44& 0.326& 1.2690&0.0040\\ \midrule \midrule
GPC (ours)& \textbf{20.08}& \textbf{0.61}& \textbf{0.40}& \textbf{0.125}& \textbf{0.3650}& \textbf{0.0010}\\ \bottomrule
\end{tabular}%
\end{table}

\subsubsection{Effectiveness of coarse-to-fine TDBS} 
The third experiment evaluates the effectiveness of our sampling strategy and coarse-to-fine training approach, focusing on depth estimation accuracy, camera pose estimation, and training convergence speed.
As illustrated in \cref{fig:TDBS}, 
our TDBS strategy outperforms the state-of-the-art method in generating accurate depth geometry.
Specifically, full-stage TDBS excels at capturing details in nearby objects, while the coarse-to-fine TDBS consistently reconstructs depth maps with realistic details across varying distances.
Additionally, coarse-to-fine TDBS yields superior results in novel view synthesis. The effectiveness of our training strategy is further supported by quantitative results shown in \cref{tab:bleff-tdbs}, where the coarse-to-fine approach reduces average camera pose estimation errors by 73.3\% and 46\% compared to uniform and full-stage strategies, respectively.
Moreover, convergence analysis, depicted in \cref{fig:opt-time}, demonstrates that TDBS not only accelerates training convergence but also enhances pose estimation accuracy. Notably, TDBS achieves minimal error within approximately 1000 epochs—only one-tenth of the epochs previously required—while avoiding local optima and enabling continuous optimization. This highlights the robustness of our strategy in improving both training efficiency and model precision.


\subsubsection{Ablation Study}
In this section, we study the impact of different components of our algorithm and the importance of Gaussian Point Constraint (GPC). 
\newline\textbf{Different Strategies.} We consider two variants of our algorithm: TDBS and GPC. \cref{tab:ablation} illustrates the results. When the GPC component is removed, the results of novel view synthesis and pose estimation do not decrease much, but if the TDBS component is removed, the results of pose estimation have a relatively large effect. This suggests that the TDBS strategy is a determining factor in the improvement of the camera pose estimation. It is because that TDBS obtain better depth geometry, pose estimation based on better depth geometry will naturally have a smaller error. Besides, TDBS and GPC have a superimposed effect on pose estimation. 
\newline\textbf{Inter-frame Point Constraint.} In addition, we compare different inter-frame point constraint methods, including ICP, point cloud distribution, and KL divergence methods. From the quantitative results in \cref{tab:ablation-point-constraint}, along with the improved quality of the view synthesis, our proposed constraint is significantly reduced by 59\% compared to both second-best constraints in terms of the camera pose estimation error.

\section{Conclusion}
\label{sec:conclusion}

In this paper, we present a novel approach, \name{}, for joint optimization of camera poses and radiance fields.
Our approach operates the coarse depth map obtained from the pre-trained depth network for depth self-supervision and proposes a coarse-to-fine sampling strategy TDBS based on the truncated normal distribution, which improves the quality of the rendered depth and speeds up the optimization.  Furthermore, our method proposes an inter-frame point cloud constraint.
This allows us to successfully improve the accuracy of the pose estimation significantly and the quality of the novel view synthesis both indoors and outdoors with large motion changes.
We implemented and evaluated our approach on different datasets, provided comparisons to other existing methods, and supported all claims made in this paper. Our experiments illustrate the generalizability of our proposed method on different datasets.


\bibliographystyle{IEEEtran}

\bibliography{glorified,new}

\end{document}

%% file: stachnisslab-latex.tex
\usepackage{graphics}           
\usepackage{times}              
\usepackage{amsmath}            
\usepackage{amssymb}            
\usepackage{graphicx}
\usepackage{algorithm}
\usepackage[noend]{algpseudocode}
\usepackage{booktabs}
\usepackage{color}
\definecolor{instructioncolor}{rgb}{.5,.5,.5}

\usepackage[font=small]{caption}

\def\eqref#1{Eq.~(\ref{#1})}

\makeatletter
\usepackage{xspace}
\DeclareRobustCommand\onedot{\futurelet\@let@token\@onedot}
\def\@onedot{\ifx\@let@token.\else.\null\fi\xspace}

\def\etal{{et al}\onedot}
\makeatother

\usepackage{array}
\newcolumntype{L}[1]{>{\raggedright\let\newline\\\arraybackslash\hspace{0pt}}m{#1}}
\newcolumntype{C}[1]{>{\centering\let\newline\\\arraybackslash\hspace{0pt}}m{#1}}
\newcolumntype{R}[1]{>{\raggedleft\let\newline\\\arraybackslash\hspace{0pt}}m{#1}}